\documentclass[letterpaper, 10 pt, conference]{ieeeconf}
\IEEEoverridecommandlockouts 
\usepackage{paralist}
\usepackage{url}
\usepackage{caption}
\usepackage{amsmath, amssymb, amssymb, amsfonts}
\usepackage{algorithm}
\usepackage{algorithmicx}
\usepackage{algpseudocode}
\usepackage{leftidx}
\usepackage{bm, bbm}
\usepackage{mathrsfs}
\usepackage{textcomp}
\usepackage{graphicx, wrapfig, lipsum}
\usepackage[hidelinks]{hyperref}
\usepackage{mathtools}
\usepackage{multirow}
\usepackage{subcaption}
\usepackage{makecell}
\usepackage{tabularx}
\usepackage[noadjust]{cite}
\usepackage{array}
\usepackage{stfloats}
\usepackage{verbatim}
\usepackage{balance}
\usepackage{tabularray}
\usepackage{lettrine}
\usepackage{xcolor}
\usepackage{tikz}
\usetikzlibrary{arrows.meta}
\usepackage[none]{comment} 
 \usepackage{booktabs}
\usepackage{siunitx}
\usepackage{gensymb}

\usepackage{soul}

\newtheorem{problem}{Problem}

\DeclareMathOperator*{\argmaxA}{argmax}
\newcommand{\real}[1][ ]{\mathbb{R}^#1}
\newcommand{\norm}[1]{\lVert #1 \rVert}

\newcommand{\vect}[1]{\mathbf{#1}}
\newcommand{\mat}[1]{\mathbf{#1}}

\begin{document}
        \title{SOMA: Stack-of-Objects Manipulation using a Quadruped Robot}

        \author{Stergios E. Bachoumas, \textit{IEEE Student Member} and Panagiotis Artemiadis$^{*}$, \textit{IEEE Senior Member}
        }



        \maketitle

        \begin{abstract}
Executing heavy-duty physical manipulation in unpredictable environments remains a major challenge for quadrupedal robots lacking dedicated manipulators. This paper presents SOMA, a hierarchical reinforcement learning framework for whole-body non-prehensile manipulation of multi-object stacks using a quadruped robot. Unlike traditional methods restricted to two-dimensional planar pushing, SOMA integrates explicit robot base-height control, unlocking full three-dimensional pushing strategies. This capability enables the quadruped to manipulate complex payload stacks as a unified system while preventing relative sliding or tipping of upper layers. The architecture factorizes control into a high-level manipulation policy using a Simplified Structured State Space Sequence (S5) encoder to reason over long-horizon temporal history and predict bounded target tracking commands via Beta distributions, and a low-level locomotion policy that dynamically adapts leg stance configurations. In extensive simulation benchmarks, SOMA achieves a $90.3\%$ success rate for single-object reposing and up to $90.1\%$ for dynamic two-box stacks under stringent error margins ($0.05\text{ m}$ position error, $1\degree$ orientation error). Finally, we validate zero-shot sim-to-real transfer on a Unitree Go1 quadruped robot across diverse payload masses and geometries, demonstrating active base-height adjustments to maintain traction and balance during heavy pushing tasks.
        \end{abstract}

        \section{Introduction}

Recent work on quadrupedal locomotion has enabled significant maneuverability across diverse and complex terrains \cite{doi:10.1126/scirobotics.aau5872, kumar2021rma}, and has also enabled gymnastic-style maneuvers such as a running backflip \cite{cafe_mpc}. However, these platforms still exhibit limitations when executing tasks that require close physical interaction with the environment. Truly autonomous robotic systems must physically interact with their environment and thus act as mobile manipulators \cite{7487545}. Recently, the robotics community has increased efforts in this direction, with promising work combining locomotion and manipulation for quadruped robots, commonly referred to as loco-manipulation.

Unlike bipeds, quadrupedal platforms inherently lack dedicated end-effectors for physical interaction. 
Nevertheless, in \cite{zhu2025versatilelocomanipulationflexibleinterlimb,cheng2023legmanip,lin2024locoman,10611307}, the authors used pedipulation, i.e., manipulation using the legs of a quadruped as an on-demand manipulator, to accomplish tasks such as opening spring-loaded doors, pushing lightweight objects out of the way, doing laundry, pushing switches, and even collecting rock samples in the wild. While these strategies prove effective for versatile manipulation, most prior work \cite{cheng2023legmanip, 10611307} strictly constrains the robot's mobility by prioritizing base static stability, with the notable exceptions of \cite{lin2024locoman,zhu2025versatilelocomanipulationflexibleinterlimb}. 

To overcome this limitation, a prominent body of work has successfully integrated robotic manipulators onto quadruped bases, achieving significant progress in mobile prehensile loco-manipulation \cite{ferrolho2023rolomarobustlocomanipulationquadruped, Mittal_2022, dadiotis2025dynamic, fu2022deep, liu2024visual, 10611066, zhou2026learning, zhu2025versatilelocomanipulationflexibleinterlimb}.
Despite these advancements, dedicated manipulator-base systems face severe physical constraints when deployed in complex environments. The increased spatial footprint and payload capacity limits of mounted arms hinder operation in tightly confined spaces and restrict the maximum load the system can manipulate. This gap becomes particularly critical in search-and-rescue operations, where robots must navigate narrow passages and interact with heavy, unstable debris under shifting dynamic loads. Consequently, achieving heavy-duty physical manipulation within highly constrained, unpredictable, and hazardous environments remains an open challenge.




\begin{figure}[t!]
    \centering
    \includegraphics[width=1.0\columnwidth]{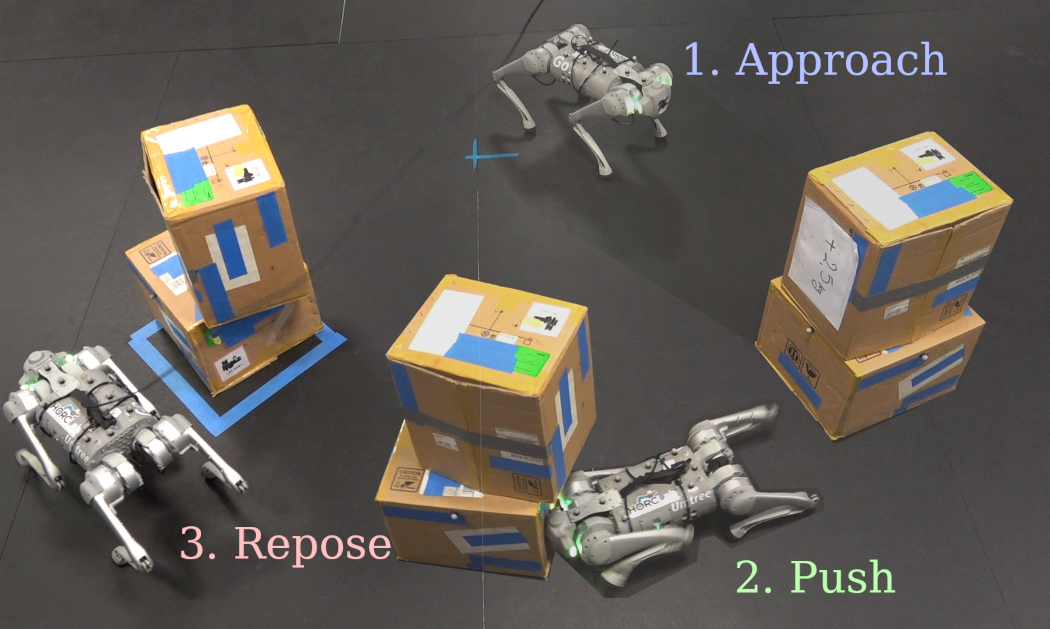}
    \caption{\textbf{Overview of SOMA}: A quadrupedal robot performing non-prehensile push manipulation on a multi-body stack of independent, unattached objects. SOMA reasons about contact height and physical parameters to push stacked payloads without destabilizing them during movement.}
    \vspace{-16px}
    \label{fig:soma}
\end{figure}
Quadrupedal robots can manipulate objects without dedicated arms either by deploying individual limbs or by leveraging their entire body for pushing interactions. While limb-based manipulation enables fine control over lightweight objects, whole-body pushing provides a compelling mechanism for transporting heavy or bulky payloads. Although non-prehensile pushing was historically underexplored \cite{latombe1991robot}, foundational work by Mason established the underlying mechanics and motion planning challenges of contact-driven manipulation \cite{mason_ijrr}. Modern optimization-based techniques have since made this problem more tractable by formulating pushing as a multi-contact planning problem \cite{graesdal2024towards}. Recently, these optimization frameworks have been extended to quadrupedal platforms to enable body-pushing behaviors \cite{rigo2022contactoptimizationnonprehensilelocomanipulation, sombolestan2023hierarchicaladaptivelocomanipulationcontrol}. However, planning through contact remains inherently difficult \cite{mason_ijrr}, and current optimization methods rely heavily on pre-specified environmental models and fixed robot parameters. As a result, existing controllers struggle to generalize across diverse or unpredictable environments, leaving generalizable, whole-body pushing locomanipulation an open challenge.


To address these limitations, we leverage learning-based paradigms \cite{11128271, jeon} and propose SOMA, a novel framework designed for whole-body object-stack manipulation on quadrupedal platforms (see Fig. \ref{fig:soma}). Built upon a hierarchical reinforcement learning architecture \cite{make4010009, jeon}, SOMA decouples control into a high-level manipulation planner that commands a low-level locomotion policy. We extend prior work by incorporating explicit control over the robot's base height, unlocking full three-dimensional pushing strategies. This vertical degrees-of-freedom integration enables the quadruped to intelligently manipulate and reconfigure complex stacks of objects from arbitrary initial states to target poses.

The contributions of this work can be summarized as follows:

\begin{itemize}
    \item A learning-based framework for quadruped whole-body manipulation of a stack of objects. Our approach incorporates direct robot base height commands, enabling full 3D pushing strategies---a crucial component missing from previous methods that restrict the robot to a predefined height and, consequently, to 2D planar pushing strategies. 
    \item The proposed policy successfully reasons about latent physical parameters of the stack---such as object mass, geometry, inter-object friction, and ground-surface friction---thereby enabling robust generalization across diverse problem instances.
    \item We evaluate the robustness of our method through extensive simulations. The proposed framework achieves a success rate of $90.3\%$ when reposing a single box-shaped object and $90.1\%$ when manipulating a double box stack, consistently meeting stringent accuracy thresholds of $1\,\degree$ in orientation error and $0.05\text{ m}$ in position error.
    \item Finally, we validate the efficacy of our proposed method in the real world by deploying it on a Unitree Go1 quadruped robot. We run an extensive suite of experiments by varying the geometry, mass, and number of objects in the stack the robot must rest on in the real world.
\end{itemize}
\section{Methods}
This work targets the following problem:

\begin{problem}[Stack of Objects]
\label{problem:stack}
A quadruped robot $R$ is tasked with pushing a stack of $N$ objects $\mathcal{S}_O=[O_1,\dots, O_N]$ in a 3D space $\mathcal{W}$ to steer the bottom-most object $B=O_1$ to a target destination while ensuring the remaining objects $\mathcal{S}_{O,n>1}=[O_2,\dots O_N]$ do not fall. Specifically, the robot must move object $B$ from its initial pose $\mathbf{X}_{B,0}=[\mathbf{R}_{B,0}, \vect{p}_{B,0}]$ to a final goal pose $\mathbf{X}_{G}=[\mathbf{R}_G, \vect{p}_G]$, where $\mathbf{R}_{B,0}, \mathbf{R}_G \in SO(3)$ represent the orientation matrices and $\vect{p}_{B,0}, \vect{p}_G \in \mathbb{R}^3$ represent the position vectors. The robot must complete the pushing and reorientation tasks within a maximum time $T_{max}$, using any part of its body. We assume that the bottom-most object is taller than the maximum height the robot can reach, that is, the height its base can reach when the legs are fully extended. Thus, to push effectively, the robot must use its entire body and contact the bottom object at different heights. Furthermore, the interaction must remain controlled; that is, the robot must avoid aggressive maneuvers and is prohibited from rolling or excessively tilting objects. 
A trial is deemed successful if the robot maneuvers the bottom object to within a position error $\tilde{d}_{terminal}$ and a yaw orientation error $\tilde{\psi}_{terminal}$ of the target goal pose without compromising the stack. Conversely, an episode terminates as a failure if the robot falls, causes any upper object to drop, or fails to achieve the target pose within the allowable error margins.
\end{problem}

We model Problem \ref{problem:stack} using the Partially Observable Markov Decision Process (POMDP) framework. Formally, a POMDP is defined as a tuple $\langle \mathcal{S}, \mathcal{A}, \mathcal{Z}, O, R, P, \gamma \rangle$.
Here, $\mathcal{S}$ is the set of states, $\mathcal{A}$ is the set of actions, $\mathcal{Z}$ is the set of observations, $O: \mathcal{S}\to \mathcal{Z}$ is a function that gives an observation from a state, $R: \mathcal{S}\times \mathcal{A}\to \real{}$ is the reward function that maps from a given state and action to a real value, $P: \mathcal{S}\times \mathcal{A}\to \mathcal{S}$ defines the distribution of next-state transitions given a state and action, and $\gamma$ is the discount factor. The learning agent's objective is to find an optimal policy (a function that maps from a given observation to a distribution over actions) $\pi^{*}:\mathcal{Z}\rightarrow \mathcal{A}$ that maximizes the expected discounted sum of rewards, also called the return:

\begin{equation}
    \pi^{*}= \argmaxA_{\pi \in \Pi}{\sum_t \mathbb{E}\big[\gamma^t r_t\big]}
\end{equation}
where $\Pi$ is the function space of all policies that have as input the observation space $\mathcal{Z}$ and as image space the action space $\mathcal{A}$, and $r_t=R(\vect{s}_{t+1},\vect{s}_t,\vect{a}_t)$ is the reward at step $t$. 

        \subsection{State and Observation Space}
Following the POMDP framework, the state of the environment at time-step $t$ is defined as $\vect{s}_t = [\vect{o}_t, \vect{n}_t]$, where $\vect{o}_t$ is the deployable/complete observation and $\vect{n}_t$ is the simulation-privileged information. To coordinate multi-stage planning, the deployable observation is further broken down into four components and is defined as $\vect{o}_t = [\vect{x}_t, \vect{g}_t, \bm{\xi}_{t}, \vect{a}_{t}]$, where $\vect{x}_t$ is the robot's proprioceptive observation, $\vect{g}_t$ is the task observation, $\bm{\xi}_t$ is the high-level command, and $\vect{a}_t$ is the low-level locomotion action. This effectively factorizes the observation space as $\mathcal{Z} \equiv \mathcal{X} \times \mathcal{G} \times \Xi \times \mathcal{A}$.
\subsubsection{Robot Proprioceptive Observation}
To capture the robot's physical state, we define $\vect{x}_t = [\bm{\omega}^\mathcal{R}, \vect{g}^{\mathcal{R}}, \vect{q}-\vect{q}_{\text{nom}}, \dot{\vect{q}}] \in \mathcal{X}\subset \real{30}$, where $\bm{\omega}^\mathcal{R} \in \real3$ is the angular velocity of the robot; $\mathbf{g}^{\mathcal{R}} \in \real3$ is the projected gravity vector in the robot frame $\mathcal{R}$; $\vect{q}, \dot{\vect{q}} \in \real{12}$ denote the robot's current leg joint positions and velocities and $\vect{q}_{\text{nom}}$$ \in \real{12}$ denotes the nominal leg joint positions that depends on the height command discussed later.

\subsubsection{Task Observation}
The task observation is defined as $\vect{g}_t = [\mat{R}_\mathcal{R}^\mathcal{W},\mat{R}_\mathcal{B}^\mathcal{R},\mathbf{p}_{B}^\mathcal{R}, \mathbf{d}_{BG}^{\mathcal{R}}, \sin{\tilde{\psi}}, \cos{\tilde{\psi}}] \in \mathcal{G}\subset \real{15}$, where $\mathbf{R}_\mathcal{R}^\mathcal{W} \in \mathbb{H}_1$ is the unit quaternion representing the orientation of the bottom box's frame $\mathcal{B}$ relative to the robot frame $\mathcal{R}$; $\mathbf{R}_\mathcal{B}^\mathcal{R} \in \mathbb{H}_1$ is the unit quaternion representing the orientation of the robot's base frame $\mathcal{R}$ relative to the world frame $\mathcal{W}$; $\mathbf{p}_{B}^\mathcal{R} \in \real{3}$ is the bottom object's position in the robot frame $\mathcal{R}$, $\mathbf{d}_{BG}^{\mathcal{R}}=(\mathbf{p}_{G}^\mathcal{R}-\mathbf{p}_{B}^{\mathcal{R}}) \in \real{2}$ is the 2D projection of the 3D vector from the box to the goal ignoring the $z$ component, and $\tilde{\psi}$ is the difference between the yaw of the box and the desired yaw at the goal position. We use the sine and cosine of the yaw error as a continuous representation of orientation \cite{Zhou_2019_CVPR}.

\subsubsection{High-Level Commands}
The high-level policy guides the low-level tracker via commands $\bm{\xi}_{t} = [\mathbf{v}_{xy}^{\text{cmd}}, \omega_{z}^{\text{cmd}}, h_{z}^{\text{cmd}}] \in \Xi \subset \real4$, where $\mathbf{v}_{xy}^{\text{cmd}} \in \real2$ represents the commanded linear base velocity in the $XY$-plane, $\omega_{z}^{\text{cmd}} \in \real{}$ is the commanded yaw rate, and $h_z^{\text{cmd}} \in \real{}$ is the target base height relative to the location of the feet, while all components of the commands are expressed in the robot's base frame $\mathcal{R}$. 

\subsubsection{Low-Level Locomotion Action}

To couple the hierarchical layers, the low-level locomotion action $\vect{a}_{t} \in \mathcal{A} \subset \mathbb{R}^{12}$ is included directly in the high-level policy's observation space at each timestep. By conditioning the high-level manipulation policy on the low-level action, the high-level controller receives continuous feedback regarding the locomotion policy's intent, allowing it to generate commands that align with executable physical dynamics.


\subsubsection{Privileged Information}
The privileged environment parameters are defined as $\vect{n}_t = [\vect{v}_{feet}, \vect{f}_c, \vect{c}, M_B, \mu, ] \in \real{30}$, where $\vect{v}_{feet} \in \real{12}$ is the vector of feet velocities, $\vect{f}_c \in \real{12}$ is the vector of feet contact forces, $\vect{c} \in \real{4}$ is the vector of contact state, $M_B$ is the total mass of the stack and $\mu$ is the box-ground friction coefficient. We use an asymmetric actor-critic structure in our framework so the critic network receives a noise-free, ground-truth observation $\hat{\vect{o}}_t$ augmented with $\vect{n}_t$ during training \cite{pinto2017asymmetricactorcriticimagebased}. The total dimension of the observation space is $dim(\mathcal{Z})=61$ for the actor and $dim(\mathcal{S})=91$ for the critic.

\subsection{Hierarchical Policy Architecture}
Each phase of the target task demands distinct planning strategies, sensing modalities, and temporal scales. For instance, the Approach Phase is a pure locomotion task governed primarily by proprioceptive feedback $\vect{x}_t$ and the stack's location. Conversely, the Pushing and Final Stages require coordinating robot ego-motion with the dynamics of the object stack, presenting a challenging whole-body manipulation problem for the combined underactuated system. 

While monolithic policies often struggle to balance such competing locomotion and manipulation objectives~\cite{jeon}, this multi-stage breakdown naturally motivates a modular design. We propose a hierarchical reinforcement learning scheme that factorizes the policy space into high- and low-level components ($\Pi \equiv \Pi_{\text{low}} \times \Pi_{\text{high}}$) mapping the previously defined observation space to execution actions.

\subsubsection{Low-Level Locomotion Policy}
The low-level policy $\pi_{\text{low}}:\mathcal{X} \times \mathcal{A}\times \Xi \rightarrow \mathcal{A}$ isolates the high-frequency locomotive aspects of the task. It maps the immediate proprioceptive observations $\vect{x}_t$, the previous joint-space residual actions $\vect{a}_{t-1} \in \real{12}$ and the tracking commands $\bm{\xi}_t$ directly to joint-space residual actions $\vect{a}_t \in \real{12}$, while all joint actions are bounded within the clipping range $[- 100,100]$. The final leg joint-level position targets $\mathbf{q}_{\text{target}} \in \real{12}$ sent to the robot's motor controllers are computed by scaling the residual action and superimposing it onto a nominal configuration baseline $\mathbf{q}_{\text{nom}}$:

\begin{equation}
    \vect{q}_{\text{target}}(t) = \vect{q}_{\text{nom}}(h_z^{\text{cmd}}) + \sigma_a \vect{a}_t
\end{equation}
where $\sigma_a=0.5$ is the action scaling factor calibrated for learning stability. The nominal baseline configuration $\mathbf{q}_{\text{nom}}(h_z^{\text{cmd}})$ adapts dynamically to height commands via a linear interpolation between a crouching configuration $\mathbf{q}_{\text{crouch}} \in \real{12}$  and a standard standing configuration $\mathbf{q}_{\text{stand}} \in \real{12}$. To evaluate this interpolation, the commanded base height $h_z^{\text{cmd}}$ is mapped to a normalized factor $\alpha \in [0, 1]$ relative to the operational height boundaries $[h_{\text{min}}, h_{\text{max}}]=[0.1,0.3]\,\text{m}$ via:

\begin{equation}
\alpha = \max\left(0, \min\left(\frac{h_z^{\text{cmd}} - h_{\text{min}}}
{h_{\text{max}} - h_{\text{min}}}, 1\right)\right)
\end{equation}
, yielding the tracking baseline:

\begin{equation}
    \mathbf{q}_{\text{nom}}(h_z^{\text{cmd}}) = (1 - \alpha)\mathbf{q}_{\text{crouch}} + \alpha\mathbf{q}_{\text{stand}}
\end{equation}

To generate the residual actions $\vect{a}_t$, $\pi_{\text{low}}$ is structured as a two-component neural network. First, an efficient Simplified Structured State Space Sequence (S5) encoder~\cite{smith2023simplified} compresses the streaming environmental observations into a compact representation while maintaining a temporal memory of the observation history. This latent representation is then processed by a Multi-Layer Perceptron (MLP) action head that outputs the mean $\boldsymbol{\mu}_t$ and standard deviation $\boldsymbol{\sigma}_t$ of a Gaussian distribution. 

During training, residual actions are stochastically sampled as $\vect{a}_t \sim \mathcal{N}(\boldsymbol{\mu}_t, \boldsymbol{\sigma}_t)$, whereas deterministic inference utilizes only the distribution mean/mode. The policy is optimized to track independent linear velocity commands $v_x^{\text{cmd}}, v_y^{\text{cmd}} \in [-1, 1]\,\text{m/s}$, angular velocity commands $\omega_{z}^{\text{cmd}} \in [-1.5, 1.5]\,\text{rad/s}$, and base height commands $h_{z}^{\text{cmd}} \in [0.1, 0.3]\,\text{m}$. The reward structure prioritizes strict command tracking while penalizing aggressive action changes, high torques, and foot slippage. The resulting lightweight locomotion policy ($\approx 100 k$ parameters) takes approximately $20$ minutes to train and is executed with a frequency of $50\,\text{Hz}$ on the real robot. The onboard joint-level PD controller tracks the output joint position targets $\vect{q}_{target}$ smoothly.

\subsubsection{High-Level Manipulation Policy}
The high-level policy $\pi_{\text{high}}:\mathcal{O}\rightarrow \Xi$ governs the long-horizon whole-body manipulation aspects of the problem. The high-level manipulation policy operates at $10\,\text{Hz}$, running at one-fifth the control frequency of the $50\,\text{Hz}$ low-level locomotion policy. At that rate, it processes the complete observation $\vect{o}_t \in \mathcal{O} \subset \real{61}$ and outputs the target locomotion command vector $\bm{\xi}_t = [\mathbf{v}_{xy}^{\text{cmd}}, \omega_{z}^{\text{cmd}}, h_{z}^{\text{cmd}}] \in \Xi \subset \real{4}$. During training of the high-level manipulation policy, the parameters of the low-level locomotion policy $\pi_{\text{low}}$ remain frozen.

Architecturally, $\pi_{\text{high}}$ consists of a sequence-modeled encoder paired with an MLP action head. Similar to the low-level architecture, a Simplified Structured State Space Sequence (S5) encoder~\cite{smith2023simplified} compresses the high-dimensional streaming observation history into a compact latent representation that preserves temporal context. To handle the bounded action space safely, the MLP actor head predicts the positive concentration parameters $\boldsymbol{\alpha}_t, \boldsymbol{\beta}_t \in [1.0, +\infty]^4$ of a multivariate Beta distribution. The softplus activation functions ensure positivity at the output. During training, a normalized command vector is stochastically sampled as $\tilde{\xi}_t \sim \text{Beta}(\boldsymbol{\alpha}_t, \boldsymbol{\beta}_t)$, where $\tilde{\xi}_t \in [0, 1]^4$. This sample is then mapped directly to the physical operational boundaries of the low-level policy via the fixed affine transformation:

\begin{equation}
    \bm{\xi}_t = \bm{\xi}_{\text{min}} + \tilde{\bm{\xi}}_t \odot (\bm{\xi}_{\text{max}} - \bm{\xi}_{\text{min}})
\end{equation}
where $\bm{\xi}_{\text{min}}, \bm{\xi}_{\text{max}}\ \in \Xi$ represent the lower and upper tracking limits of the low-level policy, respectively, and $\odot$ denotes the element-wise product. During deterministic inference, we use the mode of the distribution directly. 

Crucially, modeling $\pi_{\text{high}}$ using a Beta distribution provides a distinct mathematical advantage over standard Gaussian (Normal) formulations for bounded continuous control. While Gaussian policies during training require post-sampling artificial clipping (e.g., using a bounding function) to enforce physical limits, this clipping alters the true distribution, causing a disproportionate accumulation of probability mass at the boundaries. This boundary bias reduces policy gradients near the operational edges, causing optimization to stall. By contrast, the Beta distribution natively restricts its support to the compact interval $[0, 1]$, eliminating the need for external clipping, removing boundary bias, and ensuring stable policy gradient updates.

\subsection{Reward and Penalty Formulation}
To guide the policy through concurrent locomotion and object manipulation objectives, the reward function is defined as a weighted sum of incentive and regularizing terms:

\begin{equation}
    r_t = \sum_{i} w_i r_i(\mathbf{s}_t, \mathbf{a}_t)
\end{equation}
where $w_i$ is the weight assigned to the respective term $r_i$. Reward weights are scaled using the simulation time-step $\Delta t$ and are detailed in Table \ref{tab:incentive_weights}. 

\subsubsection{Task-Oriented Rewards}
Let $\mathbf{p}_R, \mathbf{p}_B, \mathbf{p}_G \in \real{2}$ denote the planar world positions of the robot base, box, and goal, respectively. The tracking distance between the goal and the bottom object of the stack is measured by the Euclidean norm ($L_2$) $\tilde{d}(t) = \norm{\vect{p}_G(t) - \vect{p}_B(t)}$, and $\mathbb{I}_{(\cdot)}$ represents an indicator function gating the signals.

\begin{itemize}
    \item \textbf{Base-to-Box Distance Reward} ($r_{\text{dist,rb}}$): Drives the robot toward the object using an exponential kernel based on the planar distance between the robot base and the box.
    \item \textbf{Box-to-Goal Distance Reward} ($r_{\text{dist,bg}}$): Incentivizes transporting the box toward the target position while requiring the robot to maintain active contact with the object.
    \item \textbf{Box-Goal Orientation Reward} ($r_{\text{orient,bg}}$): Minimizes the absolute yaw error $|\tilde{\psi}|$ between the box and target headings, gated to activate only when the box is within close proximity to the goal.
    \item \textbf{Directional Pushing Reward} ($r_{\text{push}}$): Promotes progress  between the box's velocity vector $\mathbf{v}_B$ and the robot base's forward heading vector $\mathbf{r}_x$. The alignment is dynamically scaled by the step-wise goal distance reduction $\delta_{BG} = \tilde{d}(t)-\tilde{d}(t-1)$, a modulation quantity that represents pushing progress.
    \item \textbf{Tiered Task Success Reward} ($r_{\text{success}}$): Provides a tiered-style incentive when the position error and orientation errors fall below predefined checkpoints. Three pairs of thresholds and checkpoint scales are used: ($\tilde{\psi}_{1}=10^\circ$, $\tilde{d}_1=0.15\ m$, $w_{\text{stage}, 1}=0.3$), ($\tilde{\psi}_{2}=7.5^\circ$, $\tilde{d}_2=0.1\ m$, $w_{\text{stage}, 2}=0.7$), and $(\tilde{\psi}_{3}=5^\circ$, $\tilde{d}_3=0.01\ m$, $w_{\text{stage}, 3}=1.0$). 
\end{itemize}
\subsubsection{Constraints and Penalties}


To promote smooth control inputs, adhere to hardware actuation limits, and preserve stack stability, we apply the following penalty terms:
\begin{itemize}
    \item \textbf{Action Rate} ($r_{\text{rate}}$): Penalizes large variations in joint commands between consecutive discrete timesteps to bound joint acceleration and ensure control smoothness.
    \item \textbf{Nominal Joint Deviation} ($r_{\text{default}}$): Penalizes the Euclidean distance between current joint positions and the default stance configuration to minimize energetic expenditures during steady-state locomotion.
    \item \textbf{Torque Expenditure} ($r_{\text{torque}}$): Minimizes motor stress by penalizing the $L_2$ norm of the commanded joint torque vector $\boldsymbol{\tau} \in \real{12}$.
    \item \textbf{Foot Slippage} ($r_{\text{slip}}$): Suppresses ground slip by penalizing the planar linear velocities $\mathbf{v}_{f, xy}$ of the feet during active stance phases.
    \item \textbf{Object Tilt} ($r_{\text{upright}}$): Enforces rigid stack stability by penalizing absolute deviations of the manipulated payload's roll $\phi_{\text{box}}$ and pitch $\theta_{\text{box}}$ angles relative to the gravity vector.
    \item \textbf{Stack Alignment Deviation} ($r_{\text{stack\_align}}$): Evaluates the horizontal geometric slip between nested layers of the multi-body object stack by tracking the maximum Chebyshev ($L_\infty$) norm of the planar position offset between any two adjacent stacked elements.
    \item \textbf{Stack Layer Slip Acceleration} ($r_{\text{stack\_acc}}$): Suppresses structural shear forces and non-rigid sliding by penalizing the maximum magnitude of the relative linear acceleration vector between the base links of adjacent stacked bodies.
\end{itemize}

\begin{table}[htbp]
    \centering
    \caption{Reward and Penalty Formulations and Weights}
    \label{tab:incentive_weights}
    \begin{tabular}{llc}
        \toprule
        \textbf{Component} & \textbf{Mathematical Formulation} & \textbf{Weight ($w$)} \\
        \midrule
        $r_{\text{dist,rb}}$      & $\exp(-\norm{\vect{p}_R - \vect{p}_B}_2 / \sigma_r)$ & $0.25$ \\
        $r_{\text{dist,bg}}$      & $\exp(-\norm{\vect{p}_G - \vect{p}_B}_2 / \sigma_g)$ & $2.0$ \\
        $r_{\text{orient,bg}}$    & $\exp(-|\tilde{\psi}|/\sigma_{\psi}) \cdot \mathbb{I}_{\norm{\vect{p}_G - \vect{p}_B} \le 0.1}$ & $10.0$ \\
        $r_{\text{push}}$         & $\exp\left( \mathbf{r}_{x} \cdot \mathbf{v}_{B} -1\right) \cdot \delta_{BG}$ & $500.0$ \\
        $r_{\text{success}}$      & $\sum_{k=1}^3 w_{\text{stage}, k} \mathbb{I}_{|\tilde{\psi}| \le |\tilde{\psi}_{k}|} \cdot \mathbb{I}_{\norm{\vect{p}_G - \vect{p}_B} \le \tilde{d}_{k}}$ & $0.1$ \\
        \midrule
        $r_{\text{rate}}$         & $\sum_{k=1}^{12} w_{\text{rate}, k} (\vect{a}_{k, t} - \vect{a}_{k, t-1})^2$ & $-0.001$ \\
        $r_{\text{default}}$      & $\sum_{k=1}^{12} w_{\text{def}, k} (\vect{q}_{k} - \vect{q}_{k}^{\text{target}})^2$ & $-0.05$ \\
        $r_{\text{torque}}$       & $\|\boldsymbol{\tau}\|_2$ & $-0.001$ \\
        $r_{\text{slip}}$         & $\sum_{f=1}^4 \|\mathbf{v}_{f, xy}\|_2^2 \cdot \mathbb{I}_{\text{contact, f}}$ & $-0.01$ \\
        $r_{\text{upright}}$      & $|\phi_{\text{box}}| + |\theta_{\text{box}}|$ & $-1.0$ \\
        $r_{\text{stack\_align}}$ & $\max_{i \in \{1, \dots, N-1\}} \|\mathbf{p}_{i, xy} - \mathbf{p}_{i-1, xy}\|_\infty$ & $-1.0$ \\
        $r_{\text{stack\_acc}}$    & $\max_{i \in \{1, \dots, N-1\}} \|\mathbf{a}_{i} - \mathbf{a}_{i-1}\|_2$ & $-0.01$ \\
        \bottomrule
        \vspace{-16px}
    \end{tabular}
\end{table}

\subsection{SOMA: Training, Domain Randomization, and Deployment}
SOMA is a full-stack framework for the manipulation of multi-object stacks using a quadruped robot. It contains the complex training pipeline, the custom domain randomization procedure, and the hardware deployment implementation. For control-system training, we use Proximal Policy Optimization (PPO) \cite{PPO} in three stages. Because we train the policies in simulation, we use domain randomization at each stage to ensure the system generalizes well to different objects and physical properties in the real world.

\subsubsection{Stage 1: Locomotion Command Tracking}
In the first stage, only the low-level locomotion policy is trained to track random locomotion commands. At the start of each episode, a random target locomotion command is generated and then resampled every $4\,\text{s}$. To make the policy robust against payloads, we sample a robot base link mass variation in the range $[-1, 3]\,\text{kg}$ from its nominal mass. To improve the policy's disturbance-rejection capabilities, we apply simulated velocity disturbances. Specifically, we sample linear velocity disturbances in the range $[-0.5, 0.5]\,\text{m/s}$ for the forward and lateral directions and $[-0.4, 0.4]\,\text{m/s}$ for the vertical direction of the robot. For the angular velocity disturbances, we sample for all main axes in the range $[-45,45]\,\degree\text{/s}$. The disturbances are applied to the base link every $1-3\,\text{s}$ \cite{pmlr-v164-rudin22a}. Finally, to facilitate sim-to-real transfer, we inject synthetic additive Gaussian noise into the locomotion policy's observation $\vect{x}_t$.

\subsubsection{Stage 2: Single-Object Manipulation}
In the second stage, we freeze the low-level policy parameters and train the high-level manipulation policy using a single object (box). The robot's task is to push the box to a goal pose with position error less than $0.05\,\text{m}$ and orientation error less than $1\degree$. To encourage policy generalization, we randomly sample each dimension of the box independently from the range $[0.35, 0.45]\,\text{m}$ at the start of each episode. We also sample the box mass uniformly from $[6, 10]\,\text{kg}$, which remains constant for the duration of the episode. Crucially, we also randomize the sliding friction coefficient between the box and the ground, sampling it uniformly from $[0.3, 1.5]$ at the start of each episode.

\subsubsection{Stage 3: Stacked-Object Manipulation}
The task is then extended to a stack of two box-shaped objects. The top object is unconstrained and free to move or slide relative to the bottom object. The robot must complete the same re-posing task as in Stage 2, but with the added requirement of keeping the stack stable, i.e., the top object must always remain on top of the bottom object.


        \section{Experiments and Results}

We evaluate our framework's capabilities in both simulation and the real world. We measure the success rate on the following task: re-posing objects with a quadruped robot under various conditions, including different object shapes and inertial parameters. We also evaluate our system's effectiveness by comparing its performance against several baseline methods in simulation.
The training process was conducted in Mjlab \cite{zakka2026mjlablightweightframeworkgpuaccelerated} on a computer equipped with a 12th Gen Intel\textsuperscript{{\textregistered}}
Core\textsuperscript{\texttrademark} i9-12900K processor featuring 24 cores, as well as an NVIDIA\textsuperscript{{\textregistered}} RTX A5500 GPU. The overall training process takes approximately $1$ hour and $20$ minutes. This is a notable reduction from the previous state of the art, which reported $15$ hours of total training time \cite{jeon}.

\begin{figure}[t!]
    \centering
    \includegraphics[width=1.0\columnwidth]{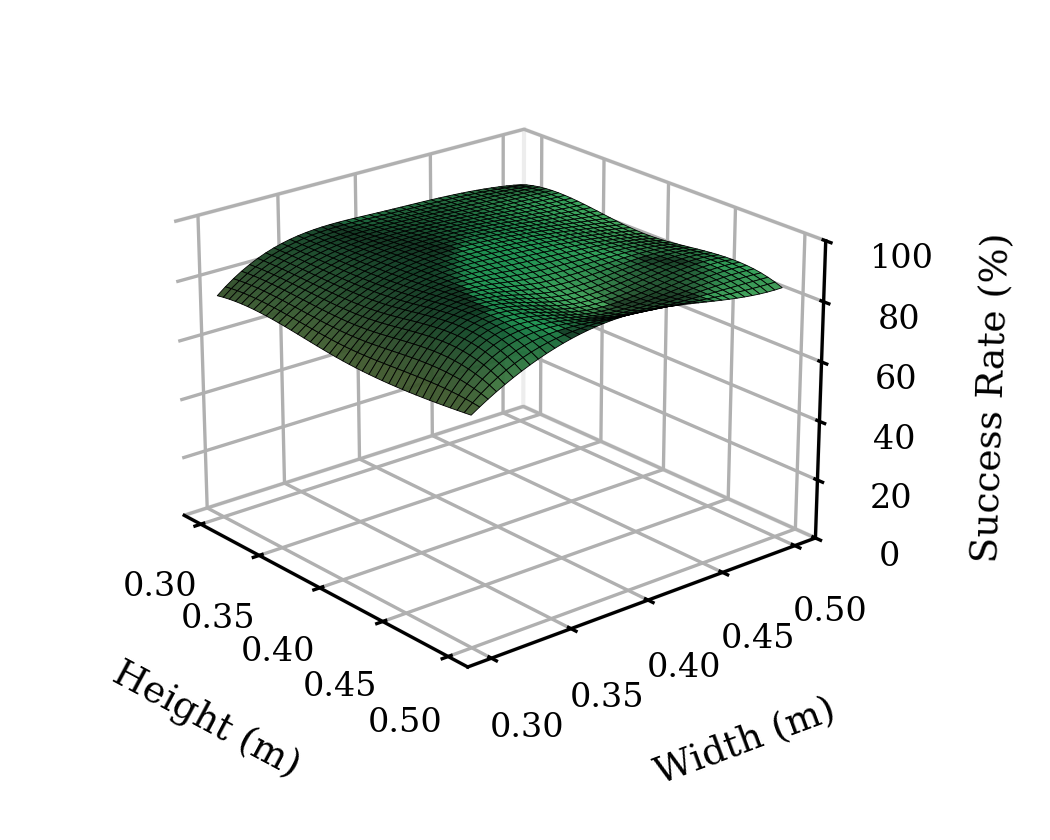}
    \caption{Manipulation success rate across varying box heights and widths for a $12\,\text{kg}$ payload. The policy maintains a $\sim 90\%$ success rate across mid-range geometries. Performance drops to $82\%$ for tall, narrow boxes prone to tipping, while wider bases sustain the $90\%$ success rate due to increased stability against overturning moments.}
    \vspace{-16px}
    \label{fig:surface_3d}
\end{figure}

\subsection{Single Box: Generalization Across Shape and Mass Variations}




To evaluate policy robustness beyond Stage~2 training bounds ($0.35\text{--}0.45\,\text{m}$ dimensions, $6\text{--}10\,\text{kg}$ mass), we systematically evaluate single-box manipulation across a discrete parameter grid of $5^3 = 125$ distinct configurations. Assuming a square base ($\text{width} = \text{depth} = w$), we sweep base side length $w \in [0.30, 0.50]\,\text{m}$ and height $h \in [0.30, 0.50]\,\text{m}$ in $\Delta w = \Delta h = 0.05\,\text{m}$ increments, and mass $m \in [4.0, 12.0]\,\text{kg}$ in $\Delta m = 2.0\,\text{kg}$ increments.

For each of the $125$ parameter combinations, we compute $500$ randomized evaluation problem instances in parallel. Each episode terminates on success, failure, or timeout, and then a new evaluation episode starts. This data collection process yielded a comprehensive suite of $72,740$ finalized evaluation episodes. During these scenarios, the box is placed in the world, and the robot's initial position is selected as a point on the perimeter of a circle $C_1$  centered at the initial position of the box with a radius $r_{C_{1}}=1.0\,\text{m}$. The goal's position is selected as another random point on the same circle $C_1$, thus setting the initial position error between the box and the goal at $1.0\,\text{m}$. The initial orientation error between the box and the goal is randomized in the range $[-180,-20]\cup[20,180]\,\degree$.


Figure~\ref{fig:surface_3d} evaluates policy performance across varying box geometries under a $12\,\text{kg}$ load. Each trial allows up to $100\,\text{s}$ to maneuver the object to within $0.1\,\text{m}$ of the goal position and $1^\circ$ of the goal orientation. Because $12\,\text{kg}$ exceeds the mass range encountered during training, this benchmark tests zero-shot generalization. Across mid-range geometries, the policy maintains a $90.3\%$ success rate. Performance drops to $82\%$ for tall, narrow geometries prone to tipping, whereas wider geometries yield $90.0\%$ success, demonstrating the policy's robustness to overturning moments.

\subsection{Efficacy of the proposed Training Strategy}
\label{sub:proposed_efficacy}


\begin{figure}[t!]
    \centering
    \includegraphics[width=1.0\columnwidth]{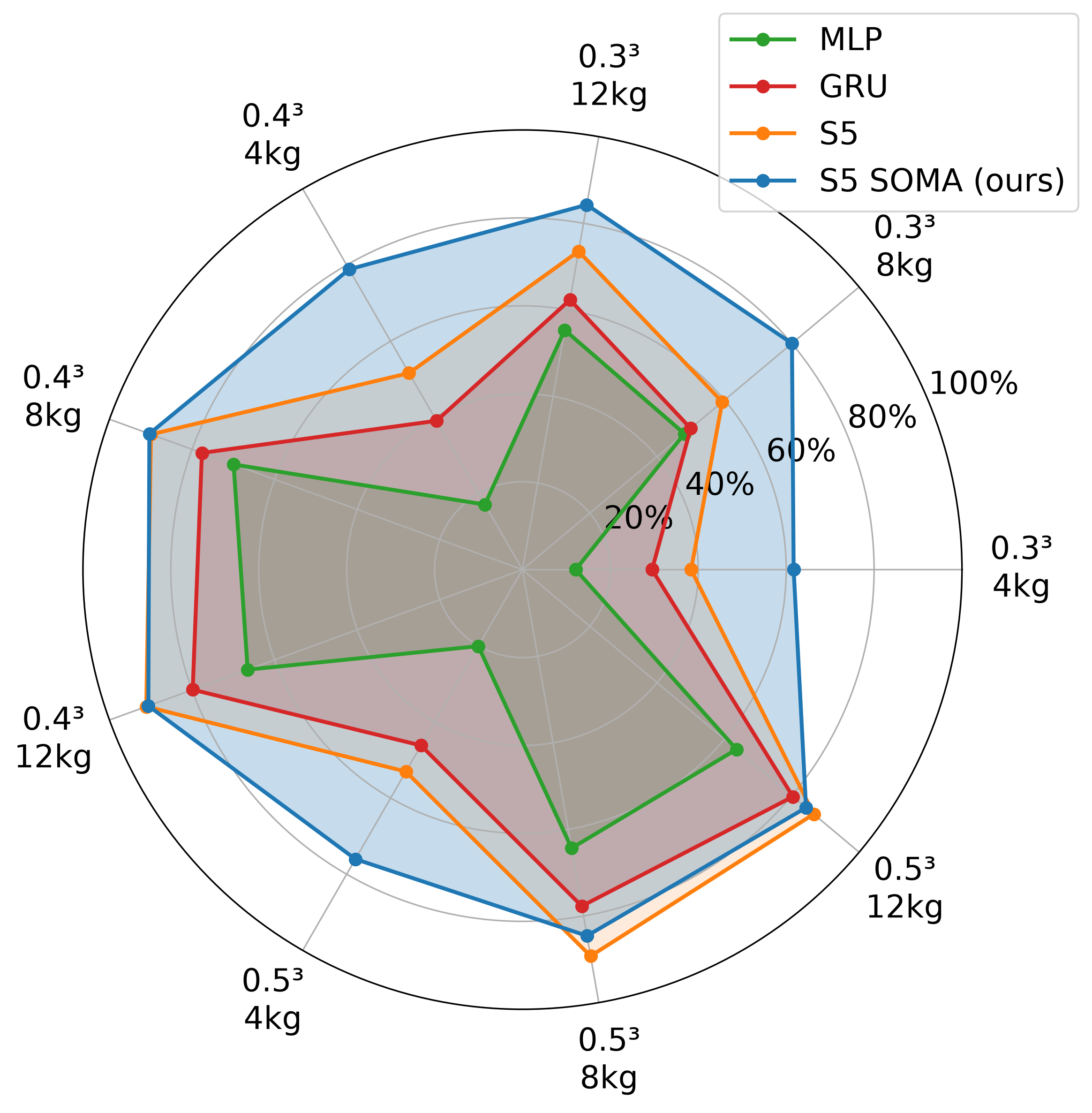}
    \caption{Manipulation success rates across varying box dimensions ($0.3^3, 0.4^3, 0.5^3\,\text{m}$) and masses ($4, 8, 12\,\text{kg}$) under strict convergence criteria ($\tilde{\psi} < 1^\circ$ heading error, $\tilde{d} < 0.05\,\text{m}$ position error). The proposed S5 SOMA framework (blue) consistently outperforms baseline models (MLP, GRU, and S5), demonstrating superior out-of-distribution generalization.}
    \vspace{-20px}
    \label{fig:radar}
\end{figure}

To evaluate both architectural choices and the proposed multi-size training strategy, we compare our model (S5 SOMA) against three baselines: a memoryless policy using only a Multi-Layer-Perceptron (MLP), a policy using a standard Gated Recurrent Unit (GRU) \cite{Chung2014EmpiricalEO} as its memory module, and a policy using a Simplified Structured State Space Sequence (S5)  memory module \cite{smith2023simplified, S5RL, bachoumas_icra_2025}. The three baseline models were trained exclusively on fixed $0.4 \times 0.4 \times 0.35\,\text{m}$ boxes with masses sampled from $[6, 10]\,\text{kg}$.

As shown in Fig.~\ref{fig:radar}, memory-augmented policies consistently outperform the memoryless MLP baseline across all conditions, demonstrating that temporal history is essential for reliable non-prehensile manipulation. Furthermore, the S5 state-space encoder systematically outperforms the GRU, highlighting its superior capacity to capture long-horizon temporal dependencies in contact-rich tasks. 

The policy with the S5 memory module outperformed the remaining baselines. To this end, we implement Stage 2 of SOMA using only the S5 model as its core component. We train a model, called S5 SOMA, on environments with boxes shaped from $0.35\times 0.35\times 0.35\,\text{m}$ to $0.45\times 0.45\times 0.45\,\text{m}$ across the same mass range [$6,10\,\text{kg}$].


Comparing training strategies, our method (S5 SOMA) yields substantial performance gains on out-of-distribution box geometries, particularly under light payloads ($4\,\text{kg}$) where single-box baselines degrade significantly. At $4\,\text{kg}$, S5 SOMA achieves success rates of $61.7\%$ ($+22.9\%$), $82.3\%$ ($+27.2\%$), and $76.0\%$ ($+22.9\%$) for $0.3^3\,\text{m}$, $0.4^3\,\text{m}$, and $0.5^3\,\text{m}$ boxes, respectively, relative to the single-box S5 baseline. At $8\,\text{kg}$, these gains persist for smaller geometries ($0.3^3\,\text{m}$: $80.0\%$ vs. $59.3\%$, $+20.7\%$), while performance remains comparable near the nominal training size ($0.4^3\,\text{m}$: $90.3\%$ vs. $89.9\%$) with a slight trade-off for larger boxes ($0.5^3\,\text{m}$: $84.6\%$ vs. $89.2\%$). At heavy boxes ($12\,\text{kg}$), both S5 variants perform within $\pm 3\%$, suggesting that mass-dominated inertial dynamics saturate controller authority regardless of geometric training distribution.


\subsection{Stack-of-Objects Evaluation}

To evaluate the framework's capacity for multi-object stack manipulation (\textit{Problem~1}), we established a three-stage experimental pipeline: environment redesign, policy fine-tuning, and downstream evaluation. First, we modified the simulation environment by adding a secondary object atop the base object to form a dynamic two-tier stack. Second, we resumed policy optimization using the top-performing single-object weights for warm-start initialization, transferring foundational pushing primitives. To promote structural stability during physical interaction, we augmented the reward formulation with penalty terms for stack alignment deviation ($r_{\text{stack\_align}}$) and inter-layer slip acceleration ($r_{\text{slip\_acc}}$).


We evaluated this fine-tuned policy across nine scenarios featuring two-cube stacks of identical dimensions. The evaluation matrix spans three cube side lengths ($0.3\,\text{m}$, $0.4\,\text{m}$, and $0.5\,\text{m}$) and three base-object masses ($4\,\text{kg}$, $6\,\text{kg}$, and $8\,\text{kg}$), while the mass of the top object was held constant at $4\,\text{kg}$. We benchmarked each configuration across $500$ parallelized simulation environments with an episode horizon of $100\,\text{s}$.

\begin{table}[t]
\centering
\caption{Termination analysis for the two stacked boxes push experiments.}
\label{tab:stack-terminations}
\resizebox{\columnwidth}{!}{%
\begin{tabular}{lcrrr}
\toprule
Side length (m) & $M_\text{bot}\ |\ M_\text{top}$ (kg) & \%Success & \%StackFail & \%Other\\
\midrule
$0.3$ & $4\ |\ 4$ & 72.4 & 12.1 & 15.5 \\
      & $6\ |\ 4$ & 71.4 & 9.1  & 23.5 \\
      & $\mathbf{8\ |\ 4}$ & \textbf{77.5} & \textbf{4.9}  & \textbf{17.6} \\
\midrule
$0.4$ & $4\ |\ 4$ & 80.1 & 7.6  & 12.3 \\
      & $6\ |\ 4$ & 87.1 & 2.7  & 10.2 \\
      & $\mathbf{8\ |\ 4}$ & \textbf{90.1} & \textbf{1.0}  & \textbf{8.9} \\
\midrule
$0.5$ & $4\ |\ 4$ & 79.9 & 8.7  & 11.4\\
      & $\mathbf{6\ |\ 4}$ & \textbf{85.3} & \textbf{0.8}  & \textbf{13.9} \\
      & $8\ |\ 4$ & 81.4 & 0.2  & 18.4 \\
\bottomrule
\end{tabular}%
}
\vspace{-0.2in}
\end{table}

As detailed in Table~\ref{tab:stack-terminations}, the framework achieved a peak task completion rate of $90.1\%$ with a $0.4\,\text{m}$ object side length and an $8\,\text{kg}$ base mass ($M_{\text{bottom}} = 8\,\text{kg}, M_{\text{top}} = 4\,\text{kg}$). Under this optimal configuration, structural stack failures dropped to $1.0\%$, with the remaining $8.9\%$ of uncompleted trials (${\%}\text{Other}$) predominantly caused by episode timeouts. 

An analysis of failure modes reveals that smaller object geometries severely compromise dynamic stack stability. Specifically, reducing the box side length from $0.5\,\text{m}$ to $0.3\,\text{m}$ caused structural collapse rates (${\%}\text{StackFail}$) to surge from a minimum of $0.2\%$ ($0.5\,\text{m}, 8\,\text{kg}$) to a maximum of $12.1\%$ ($0.3\,\text{m}, 4\,\text{kg}$) due to top-heavy tipping. A lighter base mass ($4\,\text{kg}$) further exacerbated instability across all geometries, as a heavier base lowers the stack's center of mass and provides greater resistance against disturbance torques. Although larger objects ($0.4\text{--}0.5\,\text{m}$) inherently increase resistance to tipping, overall success at $0.5\,\text{m}$ ($8\,\text{kg}$) slightly declined to $81.4\%$ due to an increase in operational timeouts ($18.4\%$). This indicates that while larger footprints improve stack integrity, they demand larger swept volumes and maneuvering room, introducing spatial execution challenges.

        \subsection{Robot Experiments}

\begin{figure*}[ht]
  \centering
  \includegraphics[width=1.0\textwidth]{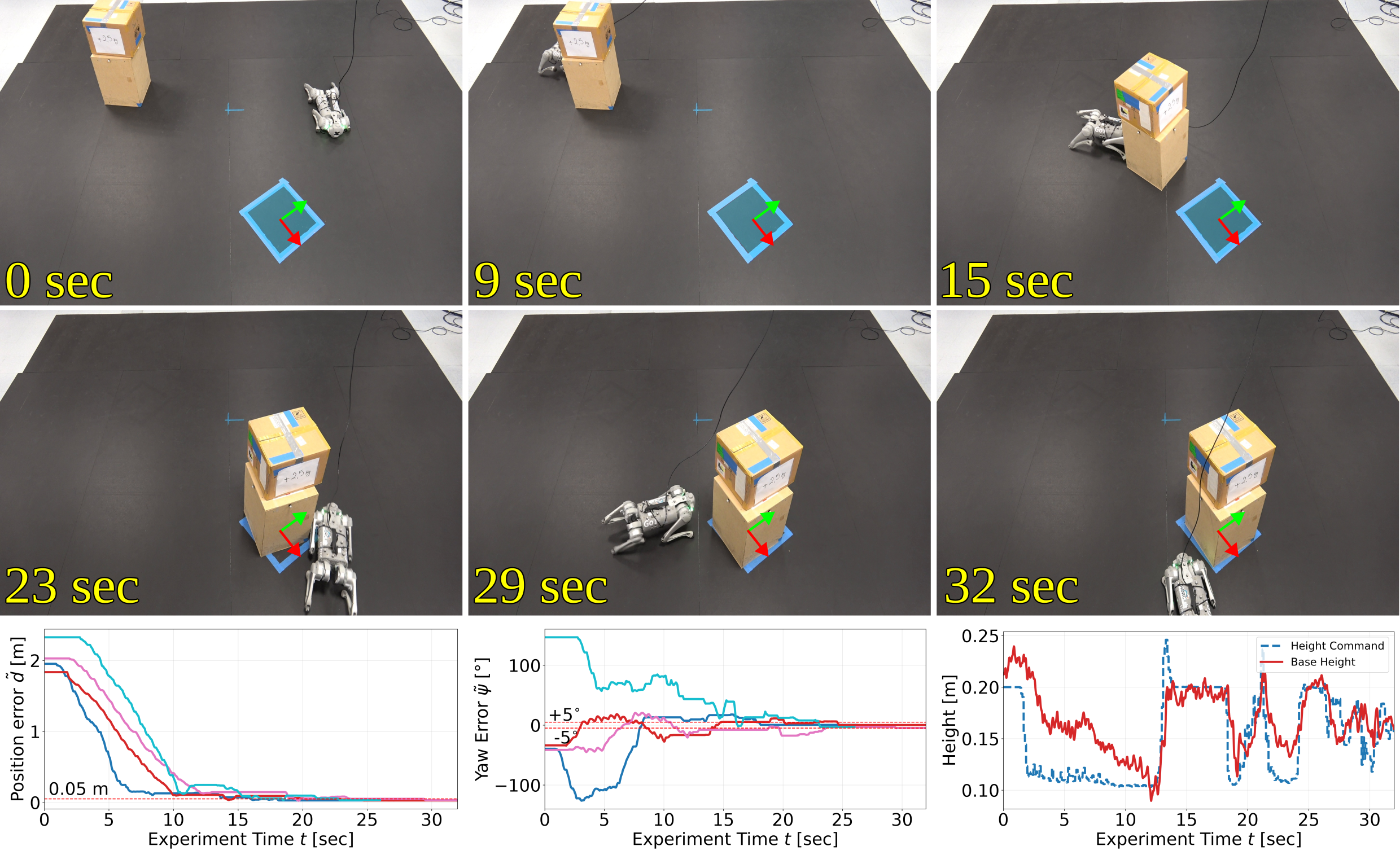}
  \caption{
  \textbf{Robot validation of SOMA double-stack loco-manipulation.} 
(\textit{Top \& Middle}): Execution sequence showing initial approach ($0\text{--}9\,\text{s}$), base-object contact alignment ($15\,\text{s}$), stack relocation ($23\,\text{s}$), fine correction ($29\,\text{s}$), and task completion ($32\,\text{s}$). 
(\textit{Bottom left}): Base-object positional error converging below $0.05\,\text{m}$ in all four trials. 
(\textit{Bottom center}): Yaw error converging within $\pm 5^\circ$. 
(\textit{Bottom right}): Active base-height lowering during contact pushing to maximize contact stability.
   }
   \vspace{-20px}
  \label{fig:robot_experiment_sequence}
\end{figure*}

\subsubsection{Experimental Setup}
Experiments were conducted on a 12-DOF Unitree Go1 quadruped robot, with the simulation-trained SOMA S5 policy deployed directly in the physical world without fine-tuning (zero-shot transfer). An eight-camera OptiTrack motion capture system (six PrimeX 13 and two PrimeX 13W cameras) tracked the robot and base object 3D poses at $240\,\text{Hz}$. Because this work focuses on learning the loco-manipulation policy rather than state estimation, we used motion capture to provide ground-truth pose measurements used throughout our framework. In the absence of external tracking infrastructure, onboard perception sensors could be integrated to localize the robot and target objects.

We evaluated five experimental configurations, varying stack geometry, total stack weight, and goal poses (detailed in Table~\ref{tab:experiment_configurations}). We considered an experiment successful if the stack did not topple and the final pose of the bottom-most object landed within a tolerance of $0.05\,\text{m}$ and $5\,\degree$ of the goal. The controller autonomously terminated the experiment when it identified that no further progress could be made. This emergent behavior significantly enhances autonomy, preventing the policy from getting trapped in a cycle of endless micro-adjustments that risk stack destabilization. 


Figure~\ref{fig:robot_experiment_sequence} (top and middle rows) depicts the temporal execution of a representative real-world trial under the \textit{Double Stack 3} configuration (Table~\ref{tab:experiment_configurations}). The robot starts $1.5\,\text{m}$ from the stack, while the base object is initialized $1.8\,\text{m}$ from the target pose with a $45^\circ$ initial heading offset (demarcated on the floor with tape as a visual reference). As shown in the bottom-left and bottom-center plots, both positional ($\tilde{{d}}$) and heading ($\tilde{\psi}$) errors decay rapidly upon initial contact. Across four independent trials, the positional error drops from about $2.0\,\text{m}$ to within the $0.05\,\text{m}$ threshold by $t \approx 18\,\text{s}$, while the yaw error converges within the required $\pm 5^\circ$ tolerance band, demonstrating high steady-state precision. Crucially, the bottom-right plot illustrates active base-height modulation during execution: while the policy maintains a nominal standing height ($\sim 0.22\,\text{m}$) during free locomotion ($0\text{--}5\,\text{s}$), it lowers the base height to $\sim 0.12\,\text{m}$ during active pushing phases ($10\text{--}20\,\text{s}$ and $25\text{--}32\,\text{s}$). This height reduction lowers the robot's center of mass and applies contact forces lower on the base object, increasing traction while preventing overturning moments that would otherwise cause the upper object to tip or slide off the stack. A video of the experiments can be found in \cite{video1}.

\begin{table}[t]
\centering
\caption{Stack object size and mass configuration for each experiment conducted.}
\label{tab:experiment_configurations}
\begin{tabular}{lcc}
\toprule
Experiment Type & Side Length ($\text{m}$) & $M_\text{bottom} \mid M_\text{top}$ ($\text{kg}$) \\
\midrule
Small Box      & $0.3 \mid -$   & $4 \mid -$ \\
Big Box        & $0.4 \mid -$   & $6 \mid -$ \\
Double Stack 1 & $0.3 \mid 0.3$ & $4 \mid 2.5$ \\
Double Stack 2 & $0.4 \mid 0.3$ & $6 \mid 2.5$ \\
Double Stack 3 & $0.4 \mid 0.3$ & $6 \mid 4$ \\
\bottomrule
\vspace{-16px}
\end{tabular}
\end{table}

        \section{Conclusion and Future Work}



In this work, we presented SOMA, a learning-based framework for quadrupedal loco-manipulation of heavy, multi-body object stacks. We validated its effectiveness through comprehensive simulation evaluations and zero-shot hardware deployment on a Unitree Go1 quadruped. Future work will extend this framework along three main vectors: integrating spatial workspace constraints for narrow-passage motion planning, replacing external tracking with onboard perception for field deployment in unstructured environments, and incorporating online parameter estimation (e.g., mass, friction, and stiffness) to dynamically tailor interaction strategies.

        \balance
        \bibliographystyle{bibliography/IEEEtran}
        \bibliography{bibliography/references}

@INPROCEEDINGS{10611307,
  author={Arm, Philip and Mittal, Mayank and Kolvenbach, Hendrik and Hutter, Marco},
  booktitle={2024 IEEE International Conference on Robotics and Automation (ICRA)}, 
  title={Pedipulate: Enabling Manipulation Skills using a Quadruped Robot’s Leg}, 
  year={2024},
  volume={},
  number={},
  pages={5717-5723},
  doi={10.1109/ICRA57147.2024.10611307}}

@ARTICLE{jeon,
  author={Jeon, Seunghun and Jung, Moonkyu and Choi, Suyoung and Kim, Beomjoon and Hwangbo, Jemin},
  journal={IEEE Robotics and Automation Letters}, 
  title={Learning Whole-Body Manipulation for Quadrupedal Robot}, 
  year={2024},
  volume={9},
  number={1},
  pages={699-706},
  doi={10.1109/LRA.2023.3335777}}

@INPROCEEDINGS{11128271,
  author={Alvarez-Padilla, Juan and Zhang, John Z. and Kwok, Sofia and Dolan, John M. and Manchester, Zachary},
  booktitle={2025 IEEE International Conference on Robotics and Automation (ICRA)}, 
  title={Real-Time Whole-Body Control of Legged Robots with Model-Predictive Path Integral Control}, 
  year={2025},
  volume={},
  number={},
  pages={14721-14727},
  doi={10.1109/ICRA55743.2025.11128271}}

@inproceedings{
smith2023simplified,
title={Simplified State Space Layers for Sequence Modeling},
author={Jimmy T.H. Smith and Andrew Warrington and Scott Linderman},
booktitle={The Eleventh International Conference on Learning Representations },
year={2023},
url={https://openreview.net/forum?id=Ai8Hw3AXqks}
}

@inproceedings{dadiotis2025dynamic,
  title={Dynamic object goal pushing with mobile manipulators through model-free constrained reinforcement learning},
  author={Dadiotis, Ioannis and Mittal, Mayank and Tsagarakis, Nikos and Hutter, Marco},
  booktitle={2025 IEEE International Conference on Robotics and Automation (ICRA)},
  pages={13363--13369},
  year={2025},
  organization={IEEE}
}

@misc{pinto2017asymmetricactorcriticimagebased,
      title={Asymmetric Actor Critic for Image-Based Robot Learning}, 
      author={Lerrel Pinto and Marcin Andrychowicz and Peter Welinder and Wojciech Zaremba and Pieter Abbeel},
      year={2017},
      eprint={1710.06542},
      archivePrefix={arXiv},
      primaryClass={cs.RO},
      url={https://arxiv.org/abs/1710.06542}, 
}

@misc{rigo2022contactoptimizationnonprehensilelocomanipulation,
      title={Contact Optimization for Non-Prehensile Loco-Manipulation via Hierarchical Model Predictive Control}, 
      author={Alberto Rigo and Yiyu Chen and Satyandra K. Gupta and Quan Nguyen},
      year={2022},
      eprint={2210.03442},
      archivePrefix={arXiv},
      primaryClass={cs.RO},
      url={https://arxiv.org/abs/2210.03442}, 
}

@misc{sombolestan2023hierarchicaladaptivelocomanipulationcontrol,
      title={Hierarchical Adaptive Loco-manipulation Control for Quadruped Robots}, 
      author={Mohsen Sombolestan and Quan Nguyen},
      year={2023},
      eprint={2209.13145},
      archivePrefix={arXiv},
      primaryClass={cs.RO},
      url={https://arxiv.org/abs/2209.13145}, 
}

@misc{ferrolho2023rolomarobustlocomanipulationquadruped,
      title={RoLoMa: Robust Loco-Manipulation for Quadruped Robots with Arms}, 
      author={Henrique Ferrolho and Vladimir Ivan and Wolfgang Merkt and Ioannis Havoutis and Sethu Vijayakumar},
      year={2023},
      eprint={2203.01446},
      archivePrefix={arXiv},
      primaryClass={cs.RO},
      url={https://arxiv.org/abs/2203.01446}, 
}

@inproceedings{Mittal_2022,
   title={Articulated Object Interaction in Unknown Scenes with Whole-Body Mobile Manipulation},
   url={http://dx.doi.org/10.1109/IROS47612.2022.9981779},
   DOI={10.1109/iros47612.2022.9981779},
   booktitle={2022 IEEE/RSJ International Conference on Intelligent Robots and Systems (IROS)},
   publisher={IEEE},
   author={Mittal, Mayank and Hoeller, David and Farshidian, Farbod and Hutter, Marco and Garg, Animesh},
   year={2022},
   month=Oct, pages={1647–1654} }

@Article{make4010009,
AUTHOR = {Hutsebaut-Buysse, Matthias and Mets, Kevin and Latré, Steven},
TITLE = {Hierarchical Reinforcement Learning: A Survey and Open Research Challenges},
JOURNAL = {Machine Learning and Knowledge Extraction},
VOLUME = {4},
YEAR = {2022},
NUMBER = {1},
PAGES = {172--221},
URL = {https://www.mdpi.com/2504-4990/4/1/9},
ISSN = {2504-4990},
DOI = {10.3390/make4010009}
}

@inproceedings{fu2022deep,
  author    = {Fu, Zipeng and Cheng, Xuxin and Pathak, Deepak},
  title     = {Deep Whole-Body Control: Learning a Unified Policy for Manipulation and Locomotion},
  booktitle = {Conference on Robot Learning ({CoRL})},
  year      = {2022},
}

@article{liu2024visual,
title={Visual Whole-Body Control for Legged Loco-Manipulation},
author={Liu, Minghuan and Chen, Zixuan and Cheng, Xuxin and Ji, Yandong and Qiu, Rizhao and Yang, Ruihan and Wang, Xiaolong},
journal={The 8th Conference on Robot Learning},
year={2024}
}

@INPROCEEDINGS{10611066,
  author={Portela, Tifanny and Margolis, Gabriel B. and Ji, Yandong and Agrawal, Pulkit},
  booktitle={2024 IEEE International Conference on Robotics and Automation (ICRA)}, 
  title={Learning Force Control for Legged Manipulation}, 
  year={2024},
  volume={},
  number={},
  pages={15366-15372},
  doi={10.1109/ICRA57147.2024.10611066}}

@ARTICLE{cafe_mpc,
  author={Li, He and Wensing, Patrick M.},
  journal={IEEE Transactions on Robotics}, 
  title={Cafe-{MPC}: A Cascaded-Fidelity Model Predictive Control Framework With Tuning-Free Whole-Body Control}, 
  year={2025},
  volume={41},
  number={},
  pages={837-856},
  doi={10.1109/TRO.2024.3504132}}

@article{
doi:10.1126/scirobotics.aau5872,
author = {Jemin Hwangbo  and Joonho Lee  and Alexey Dosovitskiy  and Dario Bellicoso  and Vassilios Tsounis  and Vladlen Koltun  and Marco Hutter },
title = {Learning agile and dynamic motor skills for legged robots},
journal = {Science Robotics},
volume = {4},
number = {26},
pages = {eaau5872},
year = {2019},
doi = {10.1126/scirobotics.aau5872},
URL = {https://www.science.org/doi/abs/10.1126/scirobotics.aau5872},
eprint = {https://www.science.org/doi/pdf/10.1126/scirobotics.aau5872}}

@article{PPO,
  title={Proximal policy optimization algorithms},
  author={Schulman, John and Wolski, Filip and Dhariwal, Prafulla and Radford, Alec and Klimov, Oleg},
  journal={arXiv preprint arXiv:1707.06347},
  year={2017}
}

@InProceedings{pmlr-v164-rudin22a,
  title = 	 {Learning to Walk in Minutes Using Massively Parallel Deep Reinforcement Learning},
  author =       {Rudin, Nikita and Hoeller, David and Reist, Philipp and Hutter, Marco},
  booktitle = 	 {Proceedings of the 5th Conference on Robot Learning},
  pages = 	 {91--100},
  year = 	 {2022},
  editor = 	 {Faust, Aleksandra and Hsu, David and Neumann, Gerhard},
  volume = 	 {164},
  series = 	 {Proceedings of Machine Learning Research},
  month = 	 {08--11 Nov},
  publisher =    {PMLR},
  url = 	 {https://proceedings.mlr.press/v164/rudin22a.html}
}

@inproceedings{kumar2021rma,
title={Rma: Rapid motor adaptation for legged robots},
author={Kumar, Ashish and Fu, Zipeng and Pathak, Deepak and Malik, Jitendra},
journal={Robotics: Science and Systems},
year={2021}
}

@INPROCEEDINGS{7487545,
  author={Rehman, Bilal Ur and Focchi, Michele and Lee, Jinoh and Dallali, Houman and Caldwell, Darwin G. and Semini, Claudio},
  booktitle={2016 IEEE International Conference on Robotics and Automation (ICRA)}, 
  title={Towards a multi-legged mobile manipulator}, 
  year={2016},
  volume={},
  number={},
  pages={3618-3624},
  doi={10.1109/ICRA.2016.7487545}}

@article{zhou2026learning,
  title={Learning Tactile-Aware Quadrupedal Loco-Manipulation Policies},
  author={Zhou, Pokuang and Zhou, Yuhao and Luu, Quan Khanh and Han, Seungho and Zhang, Heng and Huang, Binghao and Li, Yunzhu and Ajoudani, Arash and Xu, Zhengtong and She, Yu},
  journal={arXiv preprint arXiv:2604.27224},
  year={2026}
}

@inproceedings{Zhou_2019_CVPR,
title={On the Continuity of Rotation Representations in Neural Networks},
author={Zhou, Yi and Barnes, Connelly and Jingwan, Lu and Jimei, Yang and Hao, Li},
booktitle={The IEEE Conference on Computer Vision and Pattern Recognition (CVPR)},
month={June},
year={2019}
}

@misc{zakka2026mjlablightweightframeworkgpuaccelerated,
  title={mjlab: A Lightweight Framework for GPU-Accelerated Robot Learning},
  author={Kevin Zakka and Qiayuan Liao and Brent Yi and Louis Le Lay and Koushil Sreenath and Pieter Abbeel},
  year={2026},
  eprint={2601.22074},
  archivePrefix={arXiv},
  primaryClass={cs.RO},
  url={https://arxiv.org/abs/2601.22074},
}

@misc{zhu2025versatilelocomanipulationflexibleinterlimb,
      title={Versatile Loco-Manipulation through Flexible Interlimb Coordination}, 
      author={Xinghao Zhu and Yuxin Chen and Lingfeng Sun and Farzad Niroui and Simon Le Cleac'h and Jiuguang Wang and Kuan Fang},
      year={2025},
      eprint={2506.07876},
      archivePrefix={arXiv},
      primaryClass={cs.RO},
      url={https://arxiv.org/abs/2506.07876}, 
}

@article{graesdal2024towards,
  title={Towards tight convex relaxations for contact-rich manipulation},
  author={Graesdal, Bernhard Paus and Chia, Shao Yuan Chew and Marcucci, Tobia and Morozov, Savva and Amice, Alexandre and Parrilo, Pablo A and Tedrake, Russ},
  journal={arXiv preprint arXiv:2402.10312},
  year={2024}
}

@article{mason_ijrr,
author = {Matthew T. Mason},
title ={Mechanics and Planning of Manipulator Pushing Operations},

journal = {The International Journal of Robotics Research},
volume = {5},
number = {3},
pages = {53-71},
year = {1986},
doi = {10.1177/027836498600500303},

URL = { 
   https://doi.org/10.1177/027836498600500303
},
eprint = { 
    https://doi.org/10.1177/027836498600500303
}
}

@inproceedings{cheng2023legmanip,
title={Legs as Manipulator: Pushing Quadrupedal Agility Beyond Locomotion},
author={Cheng, Xuxin and Kumar, Ashish and Pathak, Deepak},
booktitle={2023 IEEE International Conference on Robotics and Automation (ICRA)},
year={2023},
}

@article{lin2024locoman,
  title={LocoMan: Advancing Versatile Quadrupedal Dexterity with Lightweight Loco-Manipulators},
  author={Lin, Changyi and Liu, Xingyu and Yang, Yuxiang and Niu, Yaru and Yu, Wenhao and Zhang, Tingnan and Tan, Jie and Boots, Byron and Zhao, Ding},
  journal={arXiv preprint arXiv:2403.18197},
  year={2024}
}

@book{latombe1991robot,
  title={Robot motion planning},
  author={Latombe, Jean-Claude},
  volume={199},
  number={1}
}

@article{Chung2014EmpiricalEO,
  title={Empirical Evaluation of Gated Recurrent Neural Networks on Sequence Modeling},
  author={Junyoung Chung and Çaglar G{\"u}lçehre and Kyunghyun Cho and Yoshua Bengio},
  journal={ArXiv},
  year={2014},
  volume={abs/1412.3555},
  url={https://api.semanticscholar.org/CorpusID:5201925}
}

@article{S5RL,
  title={Structured state space models for in-context reinforcement learning},
  author={Lu, Chris and Schroecker, Yannick and Gu, Albert and Parisotto, Emilio and Foerster, Jakob and Singh, Satinder and Behbahani, Feryal},
  journal={Advances in Neural Information Processing Systems},
  volume={36},
  year={2023}
}

@inproceedings{bachoumas_icra_2025,
  title={Learning Adversarial Policies for Swarm Leader Identification using a Probing Agent},
  author={Stergios E. Bachoumas and Panagiotis Artemiadis},
  booktitle={2025 International conference on robotics and automation (ICRA)},
  year={2025},
  organization={IEEE}
}

@online{video1,
  title        = {SOMA: Stack-of-Objects Manipulation using a Quadruped Robot},
  url          = {https://youtu.be/zm-xnCMI1uE},
  urldate      = {2026}
}
\end{document}